\documentclass[conference]{IEEEtran}
\IEEEoverridecommandlockouts

\usepackage{cite}
\usepackage{amsmath,amssymb,amsfonts}
\usepackage{algorithmic}
\usepackage{graphicx}
\usepackage{textcomp}
\usepackage{xcolor}
\usepackage{gensymb}
\usepackage{url}
\usepackage{adjustbox}
\usepackage{multirow}
\usepackage{nicematrix}
\usepackage{mwe}
\usepackage{caption}
\usepackage{subcaption}

\def\BibTeX{{\rm B\kern-.05em{\sc i\kern-.025em b}\kern-.08em
    T\kern-.1667em\lower.7ex\hbox{E}\kern-.125emX}}
\begin{document}

\title{Towards Robust On-Ramp Merging via Augmented Multimodal Reinforcement Learning
}


\author{\IEEEauthorblockN{Gaurav Bagwe\textsuperscript{1}, Jian Li\textsuperscript{2}, Xiaoyong Yuan\textsuperscript{3}, Lan Zhang\textsuperscript{1}}
\IEEEauthorblockA{\textsuperscript{1}{Department of Electrical and Computer Engineering}, Michigan Technological University, Houghton, MI, USA\\
\textsuperscript{2}School of Cyber Science and Technology, University of Science and Technology of China, Hefei, China\\
\textsuperscript{3}{College of Computing}, Michigan Technological University, Houghton, MI, USA \\
\{grbagwe, xyyuan, lanzhang\}@mtu.edu, lijian9@ustc.edu.cn}
}

\maketitle

\begin{abstract}
Despite the success of AI-enabled onboard perception, on-ramp merging has been one of the main challenges for autonomous driving. Due to limited sensing range of onboard sensors, a merging vehicle can hardly observe main road conditions and merge properly. By leveraging the wireless communications between connected and automated vehicles (CAVs), a merging CAV has potential to proactively obtain the intentions of nearby vehicles. However, CAVs can be prone to inaccurate observations, such as the noisy basic safety messages (BSM) and poor quality surveillance images. In this paper, we present a novel approach for Robust on-ramp merge of CAVs via Augmented and Multi-modal Reinforcement Learning, named by RAMRL. Specifically, we formulate the on-ramp merging problem as a Markov decision process (MDP) by taking driving safety, comfort
driving behavior, and traffic efficiency into account. To provide reliable merging maneuvers, we simultaneously leverage BSM and surveillance images for multi-modal observation, which is used to learn a policy model through proximal policy optimization (PPO). Moreover, to improve data efficiency and provide better generalization performance, we train the policy model with augmented data (e.g., noisy BSM and noisy surveillance images). Extensive experiments are conducted with Simulation of Urban MObility (SUMO) platform under two typical merging scenarios. Experimental results demonstrate the effectiveness and efficiency of our robust on-ramp merging design.

\end{abstract}
\vspace{0.5em}
\begin{IEEEkeywords}
Connected and automated vehicle, Data augmentation, Multimodal, Proximal policy optimization, Reinforcement learning, Robust on-ramp merging
\end{IEEEkeywords}
\section{Introduction}\label{sec:Introduction}
On-ramp merging has been one major bottleneck of freeway driving. Due to the reduced traffic capacity and various driving behaviors, improper handling may lead to severe traffic congestion and even accidents. According to statistics from the US Department of Transportation (USDOT), human errors are identified as the critical reason for reported road accidents-\cite{USDOT2015traffic,zhang2019machine}. Although automakers and researchers have developed and tested different levels of automated driving safety designs, such as the advanced driver assistance systems (ADAS), these systems mainly rely on onboard sensors with a limited sensing range. Hence, a merging vehicle relying on such onboard sensors can hardly observe the main road conditions automatically. In this circumstance, connected and automated vehicles (CAVs) have the potential to observe the main road driving conditions and merge properly by communicating with the roadside units (RSUs). 

Although CAVs have achieved success in many tasks, on-ramp merging is still challenging, where a merging vehicle needs to properly change speed, select a target position on the main road, and smoothly change the lane to merge. Moreover, the merging vehicle not only has to consider its own driving state but also cooperate with surrounding vehicles and predict their behaviors. Such complicated merging maneuvers require a very high sensing and path planning ability. Existing research has focused on on-ramp merging problems using algorithms from rule-based ones to machine learning (ML)-based designs, e.g., supervised learning and reinforcement learning (RL). 

Although rule-based algorithms are easy to implement, it is challenging to consider all cases that a merging vehicle may encounter. Due to the stochastic behavior of traffic flow, rule-based designs have poor generalization performance. 
In recent years, due to the excellent performance of ML in extracting the universal and discriminative features, ML-based approaches have attracted significant attention to predict driving maneuvers under complicated environments. Although supervised learning has been one of the most popular ML algorithms, it is challenging to acquire adequate labels to supervise training for complex and sequential merging behaviors, which is time-consuming and requires frequent expert human interventions. Instead, RL algorithms, especially the model-free ones, can automatically learn from the environment's observation under the supervision based on the environment's reward returns~\cite{zhao2020sim}. Thus, we use the model-free RL to automatically learn from enough observations under the challenging on-ramp merging environments. 
Specifically, we implement one popular model-free RL algorithm, the proximal policy optimization (PPO) network, because of its higher performance in continuous high-dimensional action spaces~\cite{schulman2017proximal}.


On the other hand, the training modality for on-ramp merging is another critical consideration. Most existing merging algorithms use a single-modality~\cite{triest2020learning,lin2020anti,liu2021efficient} for making decisions, which is unreliable in certain conditions~\cite{xiao2020multimodal,chaturvedi2022pay}. For example,
single-modality systems have a low accuracy when driving in adverse weather conditions, such as snow and fog~\cite{chaturvedi2022pay}. 
Thus, considering the safety-critical merging maneuvers, we propose a multi-modality design for a merging CAV by leveraging both the basic safety messages (BSM) from nearby vehicles and the images from the road-side surveillance camera. 
Moreover, to further improve the robustness of the multi-modality merging design, we introduce random amplitude scaling \cite{laskin2020reinforcement} and random Gaussian blur to augment the BSM and image observations.

By integrating all above ideas, in this paper, we present a robust on-ramp merging approach, named Robust Augmented Multi-modal and Reinforcement Learning (RAMRL). 
Specifically, RAMRL leverages the multi-modal observations from the BSM and surveillance camera to guide on-ramp merging maneuvers. The merging problem is formulated as a Markov decision process (MDP) by jointly considering safety, comfort and traffic efficiency, which is solved based on PPO networks. Augmentation techniques are implemented to further improve the robustness of RAMRL. In summary, the contributions of this paper are as follows:
\begin{itemize}
\item We propose a robust on-ramp merging approach named RAMRL to reliably handle the complicated and potential noisy driving environments. RAMRL jointly takes driving safety, comfort driving behavior, and traffic efficiency into account, which can be implemented in various merging scenarios.
   
\item 
We formulate the on-ramp merging problem as an MDP, where a multi-stage reward function is designed to guide proper merging. The MDP problem is addressed by using a model-free RL algorithm, PPO, which is further augmented by perturbing the input observations using random scaling and Gaussian blur.

\item Extensive experiments are conducted by using the simulation of Urban MObility (SUMO) platform. Experimental results show that RAMRL outperforms the baseline approaches by improving merging safety, driving comfort, and traffic efficiency. Such improvement becomes more significant under noisy environment observations, \textit{i.e.}, noisy BSM and/or poor quality surveillance camera conditions. Besides, the effectiveness of RAMRL is validated in both taper-type and parallel-type merging scenarios.

\end{itemize}

This paper is organized as follows. Section \ref{RelatedWork} reviews related work and recent progress in this topic. Section \ref{sec:ProbleStatement} formulates the on-ramp merging problem. The augmented multi-modal reinforcement learning is presented in Section IV. In Section \ref{sec:Experiments}, experiments are conducted to evaluate our design. Finally, Section VI concludes this work.
\section{Related Work}\label{RelatedWork}

On-Ramp merging problem has attracted great attention. Existing research mainly includes algorithms from rule-based ones to ML-based designs, such as supervised learning and reinforcement learning (RL). 
For rule-based on-ramp merging design, Ding \textit{et al.} \cite{Ding2020Rule} proposed a case-by-case merging strategy, which has been shown to work best for low traffic flow rates. Subraveti \textit{et al.} \cite{Subraveti2018Rule} presented another rule-based approach by creating gaps between highway vehicles without affecting the original traffic. 
However, their design is effective only at high CAV penetration and has a minimum effect on low CAV penetration. Dong \textit{et al.} proposed a predictive model based approach to predict the intentions of the vehicles using the NGSIM dataset \cite{NGSIMDataset}. The authors projected the ego vehicle on the merging lane to the highway and estimated the intentions of ramp vehicles to yield to highway vehicles. 
However, it is still difficult to accurately predict the driving behaviors of the surrounding vehicles in practice.

Wang \textit{et al.} \cite{wang2017modeling} proposed a ML-based approach by using support vector machine (SVM) to predict the merging behavior based on images from roadside cameras (RSC). Similarly, Sun. \textit{et al.} \cite{sun2018modeling} implemented an image-based logistic regression model to understand and optimize the merging policy for the merging vehicle. However, these models require a large dataset to train, and still can perform poorly in new environments. It can be infeasible to collect this huge amount of data for every new environment. Hence, RL-based methods have become popular as it uses simulated environments to ``generate" its training data.

Deep RL has been proved to achieve super human performance in complicated tasks~\cite{shao2019survey, lee2021deep, kiran2021deep}. Some recent deep DL-based research has achieved great success in on-ramp merging problems. Lin \textit{et al.} \cite{lin2020anti} proposed an end-to-end DDPG agent by using BSM data to jointly considering driving safety and comfort, where the advantages of a new driving metric for driving comfort, acceleration jerk penalty, has been validated. Lui \textit{et al.} \cite{liu2021efficient} proposed a merging strategy for multi-lane traffic using a DD-DQN model with prioritized replay based on the BSM data. This method can reduce the congestion caused by the merging vehicle as well as the average fuel consumption. Treist \textit{et al.} \cite{triest2020learning} formulated a Partially Observable MDP (POMDP) and proposes a high level decision making using Advantage Actor Critic (A2C) to control low level controllers. However, the above methods formulated RL agents using a single modality, which may generate unreliable merging maneuvers under noisy environment observations, either noisy image or noisy BSM data. Hence, in this work, a multi-modal robust RL agent is developed to automatically and reliably handle the  complicated on-ramp merging conditions, even under noisy environment observations, such as the poor networking channel or adverse weather conditions.  

\section{On-Ramp Merging Design}\label{sec:ProbleStatement}
\begin{figure}[t]
\centerline{\includegraphics[width=0.5\textwidth] {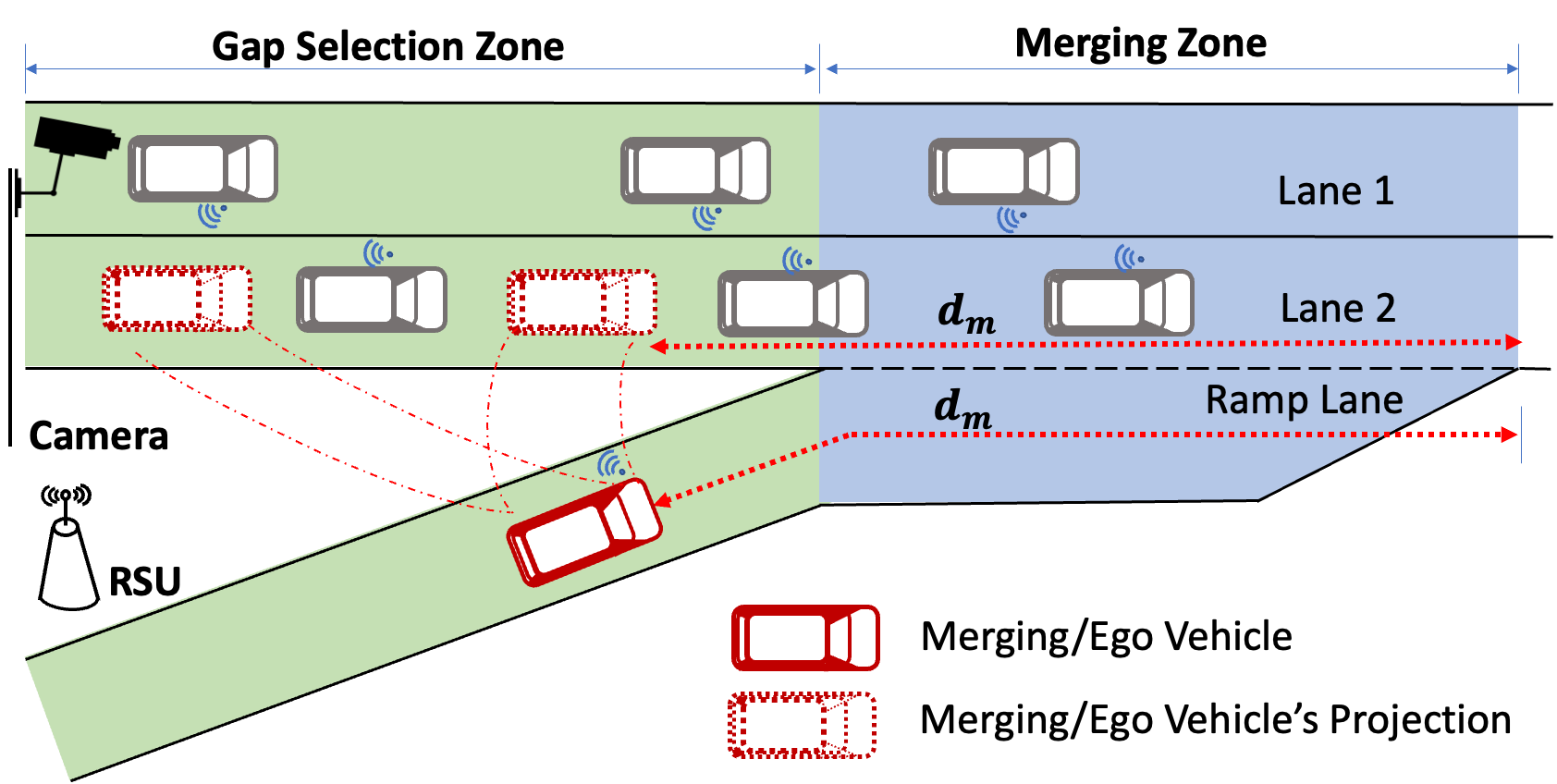}}
\caption{On-Ramp Merging Problem: General representation of merging section where the ego vehicle (in red) needs to navigate from the ramp lane to lane 2 on the highway. The scenario consists of road side unit which communicates with all the vehicles on the road and the camera which provides the input image of the merging section.}
\label{fig:mergingScenario}
\end{figure}

We consider a general on-ramp merging problem. Taking the merging topology in Fig.~\ref{fig:mergingScenario} as an example\footnote{Our approach can be generalized to different merging topologies as discussed in Section \ref{sec:training}}, a highway section consists of lane 1, lane 2, and a parallel-type ramp, where a part of the ramp is parallel to lanes 1 and 2~\cite{koepke1993ramp}. 
Each CAV has communication and computation capabilities and is equipped with an On-Board Unit (OBU) that can communicate to the roadside unit (RSU)~\cite{BSMmandate}. Besides, a surveillance camera is deployed at the roadside, as shown in Fig. \ref{fig:mergingScenario}, which captures driving images to the RSU. Consider a vehicle (red vehicle in~Fig. \ref{fig:mergingScenario}) that is trying to merge with the traffic on the highway using the ramp, named by a merging vehicle or ego vehicle. Merging with an incorrect speed and time could result in oscillation, traffic congestion, or even collisions~\cite{wang2021interpretable}. Our design aims to develop an efficient merging strategy for the ego vehicle based on multi-modal merging knowledge from the camera and RSU, to reduce traffic congestion and collision. 

Considering the erratic driving behaviors towards the ramp lane, we divide the overall ramp into two sections, a gap selection zone (marked green in~Fig. \ref{fig:mergingScenario}) and a merging zone (marked blue in~Fig. \ref{fig:mergingScenario}). When an ego vehicle enters the ramp, the RSU receives the ego vehicle’s basic safety message (BSM)~\cite{nuthalapati2016reliability} information, \textit{e.g.}, location, speed, and acceleration. Thus, the distance between the ego vehicle and the end of the merging zone $d_m$ is calculated. Meanwhile, the RSU collects observations from all the vehicles on the highway. By projecting the position of the ego vehicle from the ramp lane to lane 2 with  $d_m$, we obtain the projection of the ego vehicle as shown in Fig.~\ref{fig:mergingScenario}. We define the vehicle immediately in front of the projected ego vehicle as the first preceding vehicle and the vehicle behind as the first following vehicle. If the projected vehicle fits the gap between the first following and the first preceding vehicle, the ego vehicle selects this gap to merge. Meanwhile, the ego vehicle is given target acceleration as the input to maintain this gap. When the ego vehicle reaches the merging zone, a combination of target acceleration and steering angle is given as an input to perform the merging maneuver.

We formulate this merging problem as a Markov Decision Process (MDP). At each time step $t$, the agent, \textit{i.e.}, the ego vehicle, observes a state and takes an action. This action is implemented in the environment, which returns the next state and a reward. The following sub-sections describe the states, actions, and rewards in detail.
 
 \subsection{State Space}
 The highway section is shown in ~Fig. \ref{fig:mergingScenario}. For successful merging, an ego vehicle needs information on its own state and the states of other vehicles on the highway~\cite{nishitani2020deep}. Consider an ego vehicle at time step $t$ travels with velocity $v_{ego}$ and acceleration of $acc_{ego}$ at position $p_{ego}$. Hence, the state of the ego vehicle is defined as $S_{ego} = [v_{ego}, p_x ,p_y, acc_{ego}]$, where $p_x$ and $p_y$ are $x$ and $y$ coordinates of the ego vehicle's position. Vehicles on the highway upload their information to the RSU through BSM. The BSM consists of the current location and velocity of the vehicles~\cite{nuthalapati2016reliability,cowlagi2021Risk}. Similarly, we define the state information of the other vehicles in lanes 1 and 2. Besides, the roadside surveillance camera can capture the traffic conditions that inherently incorporate vehicles' driving behaviors and the highway traffic topology. The surveillance image will be sent to the RSU and then delivered to the ego vehicle~\cite{wen2019high}. Hence, the state of vehicles on the highway is defined as a dictionary object by, 
\begin{equation}
    S = \{ S_{ego}, S_1, S_2 , img \},
\end{equation}
where $S_{ego}, S_1, S_2 $ are the states of ego vehicle, and vehicles in lane 1 and lane 2 respectively. In each sub-state, vehicle velocity $v$, position $p_x, p_y$, and acceleration $acc$ are recorded. Additionally, the $img$ is the surveillance image from the RSU.
Thus the state information consists of two parts, BSM data $S_{ego}, S_1, S_2 \in \mathbb{R}^4$ and image data $img \in [0,255]^{W \times H \times C}$, where $W, H, C $ are the width, height and number of channels respectively.


\subsection{Action Space}
The agent is the entity which is controlled by the RL policy. In our formulation the RL controlled ego vehicle takes actions when it encounters a state observation. When the ego vehicle is located at the gap selection zone, the action space $A$ consists of the acceleration $acc_{target}$ within a range defined by $[acc_{min}, acc_{max}] $~\cite{lin2020anti}. When the vehicle is in the merging zone, the action space $A$ consists of both the acceleration $acc_{target}$ and the steering angle $\theta_{target}$ within the range $[\theta_{min}, \theta_{max}]$, where negative $\theta$ is anticlockwise steering movement and positive $\theta$ is clockwise steering movement. The action space $A$ for the ego vehicle is defined as
\begin{equation}
    A= 
        \begin{cases}
    [acc_{target}], & p_{ego} < p_{ms} \\
    [acc_{target},\theta_{target}],             & p_{ego} \ge p_{ms},
\end{cases}
\end{equation}
where $p_{ms}$ is the position at the start of merging zone and $p_{ego}$ is the ego vehicle's position. 

\subsection{Reward}
When the ego vehicle takes an action given a state, the performance of the action is evaluated based on a reward function. The reward is a type of feedback given to the agent to inform if the action is favorable. Hence, giving a multistage reward provides better feedback to the complicated merging environment~\cite{yang2019deep}. We devise a two-stage reward depending on the ego vehicle's position. Specifically, when an ego vehicle enters the ramp within the ramp lane, a projection of the ego vehicle is created. This projection will be replaced by the ego vehicle once it merges into lane 2. The two-stage reward $r_t$ at time $t$ is defined as
\begin{equation}
    r_t= 
        \begin{cases}
            r_1, & p_{ego} < p_{ms} \\
            r_2,             & p_{ego} \ge p_{ms}.
        \end{cases}
\end{equation}

Our on-ramp merging design takes driving safety, comfort driving behavior, and traffic efficiency into account. Hence, a successful merge is when the ego vehicle merges while maintaining a safe distance from surrounding vehicles and results in low traffic congestion.
When the vehicle is on the ramp and behind the merging start position, i.e., $p_{ego} < p_{ms}$, the reward is defined by  
\begin{equation}
    r_1 = r_d + r_j + r_{time},
\end{equation}
where $r_d$, $r_j$, and $r_{time}$ represents the safety reward, comfort reward, and merge time reward, respectively. Specifically, we define the safety reward $r_d$ by
\begin{equation} \label{eq:4}
    r_d =  2  - {(\frac{d_f}{d_{max}})^{-\alpha_f} }  +  {(\frac{d_r}{d_{max}})^{-\alpha_r}},
\end{equation}
where $d_f$ is the distance between the projected ego vehicle and the first following vehicle, and $d_r$ is the distance between the projected ego vehicle and the first preceding vehicle. Besides, the comfort reward $r_j$ is a to reduce the acceleration jerk/jolt which is the rate of change of acceleration, which ensures a smooth driving behavior \cite{lin2020anti}. We define $r_j$ by 
    \begin{equation} \label{eq:5}
        r_j = - \alpha_j \times \frac{|\dot{acc_{ego}}|}{j_{max}},
    \end{equation}
where $\dot{acc_{ego}}$ is the rate of change of acceleration, $j_{max}$ is the maximum allowed acceleration jerk used to normalize the reward between 0 and 1. This rewards acts as a penalty to encourage smooth driving behavior of the ego vehicle. When the ego vehicle does not maintain a constant acceleration it would receive a penalty. Intuitively, this penalty has to be low else the amount of collisions by the ego vehicle may be high. For example, when the accumulated rewards by acceleration jerk and the success reward may be less than the accumulated rewards by the acceleration jerk and collision the ego vehicle may choose to collide~\cite{lin2020anti}.
We consider $r_{time}$ as another penalty to reduce the merging time. Similar to the acceleration jerk reward, this value should also be low to reduce the collisions. Besides, this reward considers velocity to encourage lateral motion. Hence, the merge time reward $r_{time}$ is defined as
\begin{equation} \label{eq:6}
    r_{time} = - \alpha_{time} \times time_{ego} + \alpha_v * v_{ego},
\end{equation}
where  $\alpha_{time}$ and $\alpha_v$  are weights for merge time and velocity rewards respectively, $time_{ego}$ is the time since the ego vehicle entered the ramp, and $v_{ego}$ is ego vehicle's velocity.

When the ego vehicle is in the merging section, \textit{i.e.}, ahead of the merging position $p_{ego} \ge p_{ms}$, the projected ego vehicle is replaced with the actual ego vehicle. The reward at this stage $r_2$ is defined as
\begin{equation}
    r_2 = r_d + r_j + r_o + r_c + r_{end} + r_{time}, 
\end{equation}
 where $r_d$, $r_j$, and $r_{time}$ are the safety reward, comfort reward, and merge time reward as defined above in equation~\ref{eq:4}, \ref{eq:5}  and \ref{eq:6}, respectively. Additionally, $r_o$, $r_c$ and $r_{end}$ represent the smooth merging reward, congestion reward and terminal reward, respectively. Specifically, $r_o$ measures the smoothness of the merging behaviour~\cite{hu2020end}, which can be given by 
    \begin{equation}
        r_o = - \alpha_o  \times \dot{\theta_o},
    \end{equation}where $\alpha_o$ is the reward weight coefficient and $\dot{\theta_o} $  is the rate of change of steering angle. Besides, the congestion reward $r_c$ is the reward to reduce traffic congestion due to on-ramp merging of the ego vehicle. We define $r_c$ as
    \begin{equation}\label{eq:rewCongestion}
        r_c = \alpha_{s_{l2}} \times \frac{\overline{v}_{lane2}}{v_{max}} + \alpha_{s_{l1}} \times \frac{\overline{v}_{lane1}}{v_{max}},
    \end{equation}
 where $\overline{v}_{lane2}$ and $\overline{v}_{lane2}$ are the average velocity of $M$ vehicles and $N$ vehicles in lane 2 and lane 1, respectively; $v_{max}$ is the maximum allowed velocity used to normalize the reward between 0 and 1. The terminal reward $r_t$ is the reward given at the end of the episode. A termination can happen due to collision or when the ego vehicle successfully reaches the end of the merging section. Hence, the terminal reward $r_{end}$ is defined as
    \begin{equation}\label{eq:rewSucCollision}
    r_{end}= 
        \begin{cases}
    -a_{collision}, & collision =1 \\
    +a_{success}, & p_{ego} \ge p_{es} \\
    0,             & else,
    \end{cases}
    \end{equation} where the $a_{collision}$ is the collision coefficient given to the ego vehicle when it collides with the surrounding vehicles. $a_{success}$ is the success coefficient given when the ego vehicle successfully reaches the end of merging section.


\section{Augmented Multi-modal Reinforcement Learning}\label{sec:RLPreliminaries}
Reinforcement learning (RL) has been a popular learning approach to automatically solve MDP problems. With every action that the agent performs, the environment returns the rewards. Based on these rewards, the RL algorithm updates the policy of the agent. This process is continued until the agent learns to take correct actions by maximizing the expected rewards. 

We use the model-free policy-based RL algorithms as they are suitable for continuous action spaces and have better convergence~\cite{sutton2018reinforcement,deepRL-2020}. Schulman \textit{et al.} introduced a model-free policy optimization method called Trust Region Policy Optimization (TRPO) \cite{schulman2015trust}. TRPO updates the policy with a constraint on the size of the update \cite{schulman2017proximal}. The surrogate objective function is defined as:
\begin{align}
    \underset{\theta}{maximize} \ L_{\theta_{old}} \left( \theta \right) \\
    s.t. \ D_{KL}^{max} \left( \theta_{old},\theta)  \right) \leq \delta,
\end{align}
where $\theta$ is the network parameters for the policy $\pi_{\theta} \left( a|s) \right)$, and $\theta_{old}$ are the parameters of the old policy. This policy is updated on the trust region constraint of  KL divergence denoted by $D_{KL}$, where $\delta$ is the bound on KL divergence~\cite{schulman2015trust}.

The proximal policy optimization (PPO) algorithm uses first-order derivative, which makes as reliable and efficient as the TRPO while being simpler to implement~\cite{schulman2017proximal}. Specifically, PPO uses a clipped surrogate objective which is a modification of the objective of the TRPO network. The network is trained to maximize the expected returns from the loss function~\cite{schulman2017proximal}, which can be given by
\begin{equation} \label{eq:MainPPO_loss}
\begin{split}
    &L_t^{CLIP + VF + S} ( \theta ) =  \\
    &\hat{\mathbb{E}_t} \left[ L_t^{CLIP} ( \theta ) - c_1 L_t^{VF} ( \theta )  + c_2 S_{\left[ \pi_{\theta} ( s_t ) \right]} \right],
\end{split}
\end{equation}
where $c_1$ and $c_2$ are coefficients and $S$ denotes the entropy bonus. $L_t^{CLIP}$ is the clipped objective, the second term $L_t^{VF} \left( \theta \right) $ is the Mean Squared error between the value function and the target value function, and the third term gives entropy bonus. The individual losses are defined as
\begin{equation}\label{eq:PPO_loss}
\begin{split}
    L_t^{CLIP} ( & \theta )  = \\
     &\hat{\mathbb{E}_t} \left[ min( r_t(\theta)  \hat{A_t},  clip( r_t(\theta) , 1- \epsilon, 1+ \epsilon ) )\right],
\end{split}
\end{equation}
and
\begin{equation}
    r_t(\theta) = \frac{\pi_{\theta}\left(a_t, s_t\right)}{\pi_{\theta_{old}}(a_t, s_t)},
\end{equation}
where $\epsilon$ is a clipping coefficient and $\hat{A_t}$ is the advantage function \cite{schulman2017proximal}. Finally, we update the policy using equation~(\ref{eq:MainPPO_loss}). 

Additionally, to achieve better robustness, we further augment the state information by introducing the random amplitude scaling~\cite{laskin2020reinforcement} and random Gaussian blur. Specifically, Raffin \textit{et al.}~\cite{raffin2021stable} analyzed the benefits of augmenting the data for RL algorithms, which becomes essential for the safety-critical merging scenarios. Along that line, we first augment the BSM data by adding Gaussian noise to the source information. The velocity, for example is augmented as follows, 
\begin{equation}\label{eq:augNoise}
    v_{aug} = v_{obs} + \mathbb{G(\mu,\sigma)},
\end{equation}
where $v_{obs}, v_{aug}$ is the observed and augmented velocity respectively, and $\mathbb{G(\mu,\sigma)}$ is the Gaussian noise with mean $\mu$ and standard deviation $\sigma$. Similarly, we augment the position and acceleration. Additionally, uniform noise is used as the standard deviation to add noise to the surveillance images.



\section{Experiments}\label{sec:Experiments}
This section first presents the experimental setup, including the on-ramp merging simulator, on-ramp merging topology, the RL framework, and the baseline approaches. Then, we evaluate the robustness of the proposed approach versus the traditional approaches. Finally, we consider different ramp topologies to validate the generalization of RAMRL.

\subsection{Experimental Setup}\label{subsec:ExpSetup}
In the following, we will describe the experimental setup in detail. Specifically, we discuss how we modify the simulator to the on-ramp merging environment. Additionally, we define the parameters for training the RL policy. Besides, the interactions between the RL policy and the driving simulation environments will be introduced.

\textbf{On-Ramp Merging Simulator.}
We develop an on-ramp merging simulator using Simulation of Urban MObility (SUMO) platform~\cite{lopez2018microscopic}. SUMO is an open-source microscopic traffic simulator consisting of rule-based car-following models to simulate real-world driving conditions. These car-following models can be controlled online using Traffic Control Interface (TraCI). TraCI is an interface between SUMO and python, which allows information retrieval and manipulations of objects such as vehicles, and traffic lights. Using TraCI, SUMO retrieves information about the current state of the vehicles and controls the speed and driving behavior of the ego vehicle in real-time~\cite{wu2017flow}. 

To create the environment, we build on the sumo-rl environment~\cite{Lucas2019sumorl}, which is primarily built for traffic signal control. We modify this environment for on-ramp merging tasks. 
SUMO is controlled by Traffic Control Interface (TraCI), which acts as an online controller to control SUMO for the RL framework. Using TraCI, we also get the BSM information of the non-RL vehicles. TraCI passes the state information from SUMO to the RL framework and executes the action from the RL framework to create the next state. 

\begin{figure}[!tb]
\centerline{\includegraphics[width=0.5\textwidth]{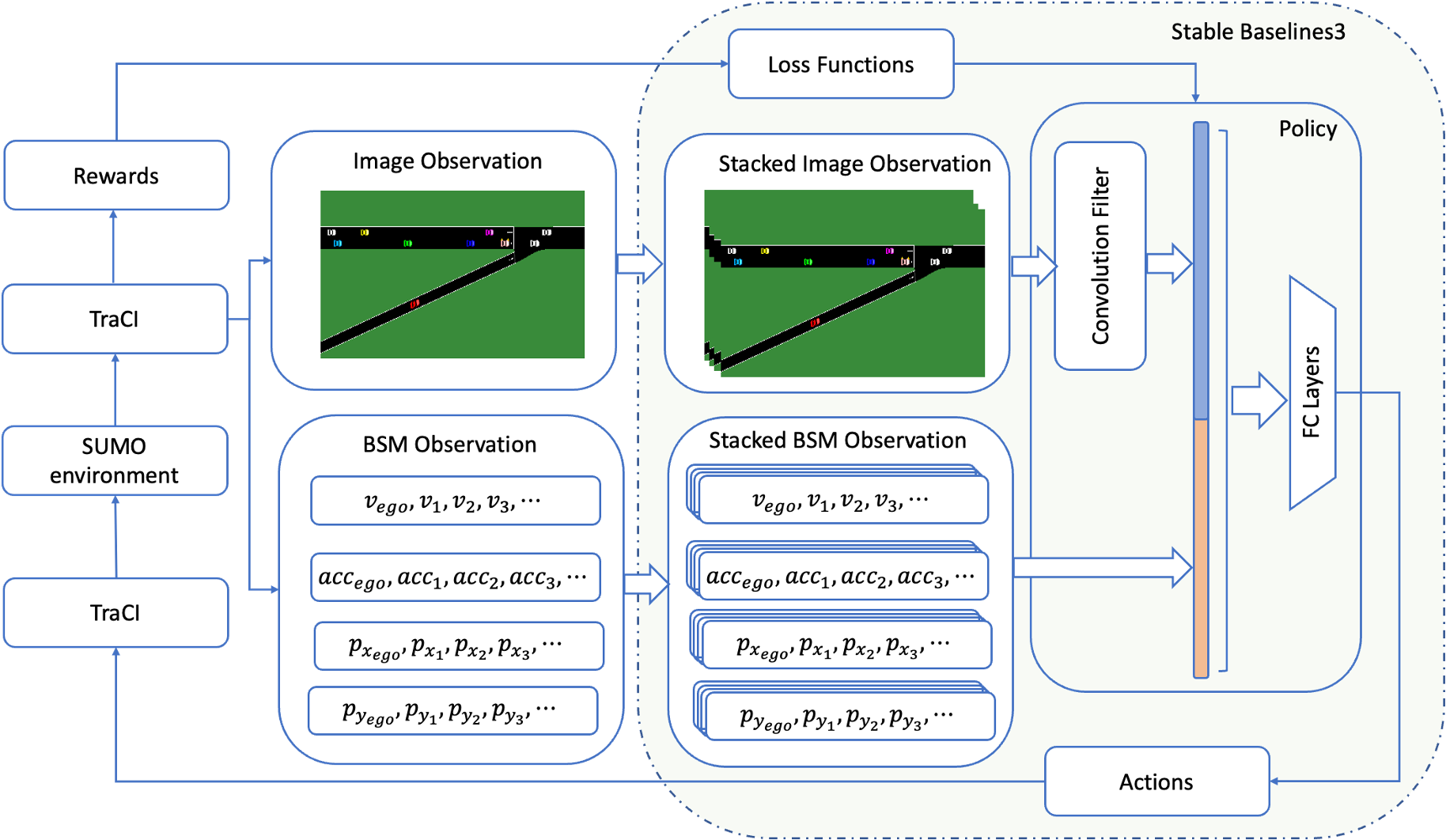}}
\caption{Experimental Testbed Setup. The state information is extracted from the SUMO environment, consisting of the image and BSM data. TraCI transmits this information to the Stable-Baselines3 framework. Consecutive frames of the image information and BSM data forms the observation which is given to the PPO policy. The PPO policy predicts an action from the observations which is given to TraCI. TraCI takes performs the action on rl vehicle and returns the next state and rewards. The rewards are used to calculate the loss which updates the policy function.
}
\label{fig:SumoRamp}
\vspace{-1em}
\end{figure}

\begin{figure}[!tb]
     \centering
     \begin{subfigure}[b]{0.5\textwidth}
         \centering
         \includegraphics[width=\textwidth] {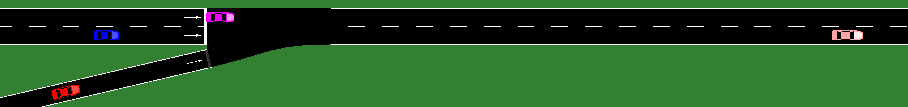}
         \caption{Taper type topology}
         \label{fig:tapertop}
     \end{subfigure}
     \begin{subfigure}[b]{0.5\textwidth}
         \centering
         \includegraphics[width=\textwidth] {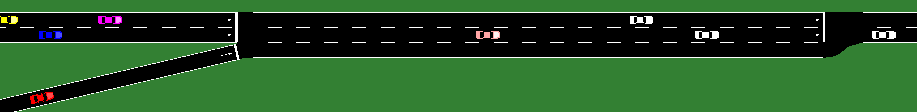}
         \caption{Parallel type topology}
         \label{fig:paralleltop}
     \end{subfigure}
     \caption{Different ramp topologies. Fig. \ref{fig:tapertop} is an example of the taper type topology, where the ego vehicle does not have an acceleration lane before the merge intersection. Fig.~\ref{fig:paralleltop} is an example of a parallel type design where the ego vehicle has a additional lane parallel to the main road before the merging intersection.}
     \label{fig:RampToplogies}
     \vspace{-1em}
\end{figure}

\textbf{On-Ramp Merging Topology.}
We design the taper road network in SUMO with the following specifications as shown in Fig.~\ref{fig:tapertop}. The main road and ramp section length is set to $300$m pre-intersection and $100$m post-intersection. Additionally, Krauss car-following model \cite{sun2018modeling} controls the non-RL vehicles on the main road with a maximum velocity of $13.89m/s$. Finally, the traffic flow is defined by the number of vehicles per hour, which is set to 1440 vehicles per hour per lane. Similarly, we design a parallel type design shown in Fig. ~\ref{fig:paralleltop}. The parallel type design has a third lane, which runs parallel to lanes 1 and 2, of the length of $100$m to accelerate the speed of the highway. The pre-intersection, post-intersection length,  maximum velocity, and the number of vehicles are equivalent to that of the taper-type design.

\vspace{0.2em}
\textbf{State Observations.}
During the training stage, as shown in Fig. \ref{fig:SumoRamp}, an ego vehicle in red is spawned on the ramp, and Krauss controlled non-RL vehicles are spawned on the main highway. Each episode runs with one ego vehicle, and the episode terminates when either the vehicle successfully merges or collides with another vehicle. The state contains a top-down image of the road segment, ego vehicle, a tuple of 4 vehicles on lane 2 and 2 vehicles on lane1. Hence, the state information is defined as 
\vspace{-0.3em}
\begin{equation}
\begin{split}
    S = \{&(v_{ego},v_1,v_2, v_3, v_4, v_5, v_6 ), \\
               &(acc_{ego},acc_1,acc_2, acc_3, acc_4, acc_5, acc_6 ),\\
               &(p_{x_{ego}},p_{x_{1}},p_{x_{2}}, p_{x_3}, p_{x_4}, p_{x_5},p_{x_6} ),\\
               &(p_{y_{ego}},p_{y_1},p_{y_2}, p_{y_3}, p_{y_4}, p_{y_5},p_{y_6}),\\
               &img_{[W \times H \times C]}
              \},
\end{split}
\end{equation}
where $v_i,acc_i, p_{x_i}, p_{y_i}$ are the velocities, accelerations positions of the vehicles, $i \in \{ego, 1,2,3,4,5,6\}$. $img_{[W \times H \times C]}$ denotes the image observation with width $W$, height $H$ and channel $C$. 

\vspace{0.2em}
\textbf{Reinforcement Learning Framework.}
We implement the proposed approach using stable-baselines3~\cite{raffin2021stable}. 
We design a custom policy network, which is updated by the loss function defined by Equation~\ref{eq:PPO_loss}. Additionally, we use a gym wrapper to create a stack of four observations to simulate motion information.
The complete training architecture is illustrated in Fig.~\ref{fig:SumoRamp}. TraCI passes the image and BSM state information to the policy defined by stable-baselines3. The policy passes the predicted action to TraCI, which executes the next step in the environment. Stable-baselines3 gets the rewards from the environment and makes a policy update. Additionally, TraCI gets the next state information, and the training cycle continues. For running the experiments, we create the road network, followed by defining the non-RL vehicles and the custom policy network.

We set the parameters for the reward functions in Section ~\ref{sec:ProbleStatement}. The maximum acceleration jerk ($j_{max}$), maximum velocity $v_{max}$ is set to $2.6$ and $13.89$ respectively. Additionally, $\alpha_{f}$, $\alpha_{d}$, $\alpha_{j}$ is set to 0.5, 0.5 and 0.8 respectively. Moreover, $\alpha_{s_{l2}}$ is 1 and $\alpha_{s_{l1}}$ is 0 to consider only the congestion and efficiency of vehicles in lane 2. Finally, we set the values in Gaussian noise in Section ~\ref{sec:RLPreliminaries}. Inspired by the noise level from Zaid \cite{Zaid2019SAE}, we augment the position information with Gaussian noise $n\sim\mathcal{G}(1.5,1)$, the speed with $n\sim\mathcal{G}(0.2777,1)$ and acceleration with $n\sim\mathcal{G}(0.3,1)$. Besides, we augment the image with a Gaussian filter to simulate the inaccuracy due to weather conditions. During training, a uniform noise with a random value between $[0,1]$ is used as the standard deviation to add noise to the surveillance images.

\vspace{0.2em}
\textbf{Baseline Approaches.}
We compare our RAMRL approach with Krauss car-following model~\cite{krauss1998microscopic} and single modality models. We compare the rule-based and RL-based methods using SUMO's default Krauss car-following model. Additionally, to compare to traditional RL approaches, we compare the single modality models individually trained on perfect BSM and image observations.


\begin{figure}[!tb]
\vspace{-2em}
\centerline{\includegraphics[width=0.4\textwidth]{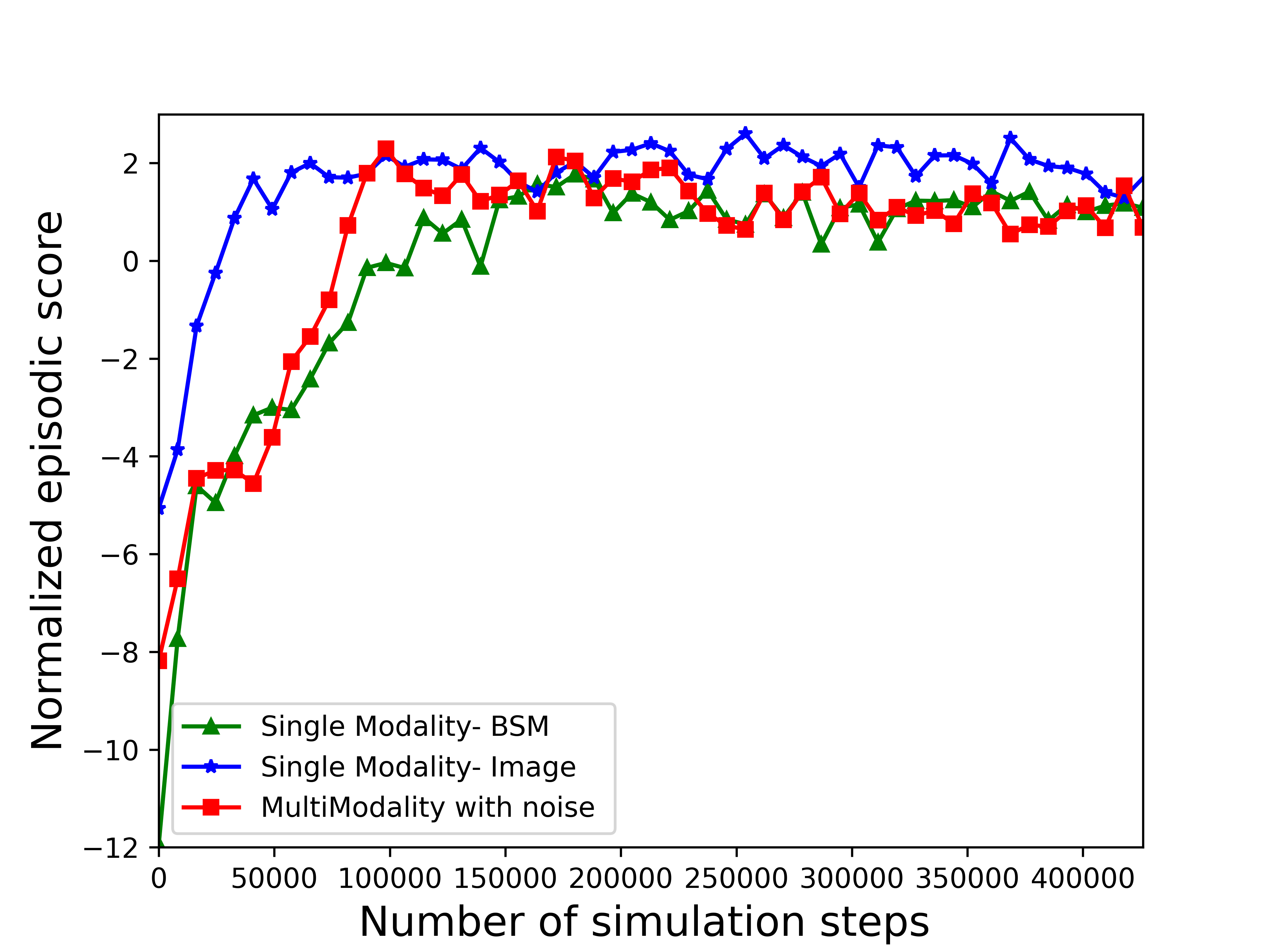}}
\caption{Learning curves for single modality and the proposed multi-modal data is augmented with noise for GSM, velocity and acceleration measurements, while the image data is augmented from surveillance images.}
\label{fig:trainingleModality}
\vspace{-1em}
\end{figure}
\begin{table*}[!tb]
\centering
\begin{tabular}{|c|c|c|c|c|c|}
\hline
Type of modality &
  \begin{tabular}[c]{@{}c@{}}Average velocity\\  $(m/s)$\end{tabular} &
  \begin{tabular}[c]{@{}c@{}}Average merge time\\ $(s)$\end{tabular} &
  \begin{tabular}[c]{@{}c@{}}Average number\\  of collisions\end{tabular} &
  \begin{tabular}[c]{@{}c@{}}Average acceleration \\ jerk $(m/s^3)$\end{tabular} &
  \begin{tabular}[c]{@{}c@{}}Normalized\\  episodic rewards\end{tabular} \\ \hline
  
Krauss car following model   & 11.43 & 105.1 & 0.00 &-0.0838 &  -16.6363 \\ \hline
Single-modality RL (BSM)          & 11.44 & 68.5  & 0.00 & -0.0440 & 2.4577   \\ \hline
Single-modality RL (image)        & 11.39 & 48    & 0.10 &  -0.1875  &1.4292   \\ \hline
Multi-modality RL (BSM+image) & 11.42 & 48.1  & 0.00 & -0.0897 &  1.7980   \\ \hline
Proposed RAMRL         & 11.72 & 55.6  & 0.00 &-0.0788 &  2.4958   \\ \hline
\end{tabular}

    \caption{Merging performance comparison 
    under perfect environment observations.}
    \label{tab:modalityTesting}
\end{table*}
\begin{table*}[!tb]

\begin{adjustbox}{width =1\textwidth}

\begin{tabular}{|c|c|c|c|c|c|c|}
\hline
Amount of noise added &
  Type of modality &
  \begin{tabular}[c]{@{}c@{}}Average velocity\\  $(m/s)$\end{tabular} &
  \begin{tabular}[c]{@{}c@{}}Average merge time\\ $(s)$\end{tabular} &
  \begin{tabular}[c]{@{}c@{}}Average number\\  of collisions\end{tabular} &
  \begin{tabular}[c]{@{}c@{}}Average acceleration \\ jerk $(m/s^3)$\end{tabular} &
  \begin{tabular}[c]{@{}c@{}}Normalized\\  episodic rewards\end{tabular} \\ \hline
\multirow{3}{*}{No noise}   & SMRL(BSM)   & 11.44 & 68.5  & 0.00 & -0.0440 & 2.4577 \\ \cline{2-7} 
                            & SMRL(image) &  11.39 & 48    & 0.10 &  -0.1875  &1.4292       \\ \cline{2-7}
                            & RAMRL & 11.72 & 55.6  & \textbf{0.00} &-0.0788 &  \textbf{2.4958} \\ \hline 
\multirow{3}{*}{10\% noise} & SMRL(BSM)   & 12.50 & 37.8 & 0.00 & -0.0152 & 2.4505 \\ \cline{2-7} 
                            & SMRL(image) & 11.59 & 43.8 & 0.20  & -0.1953         & 1.0838        \\ \cline{2-7} 
                            & RAMRL & 11.61 & 43.6 & \textbf{0.00} & -0.0791 & \textbf{2.8004} \\ \hline
\multirow{3}{*}{25\% noise} & SMRL(BSM)   & 12.25 & 38.7 & 0.20 & -0.0920 & 2.4990 \\ \cline{2-7} 
                            & SMRL(image) & 11.26    & 46.0 & 0.30
                            & -0.1656 &  0.7506      \\ \cline{2-7} 
                            & RAMRL & 12.19 & 47.6 & \textbf{0.00} & -0.1206 & \textbf{3.4590} \\ \hline
\multirow{3}{*}{50\% noise} & SMRL(BSM)   & 12.34 & 36.1 & 0.30 & -0.0119 & 1.8820 \\ \cline{2-7} 
                            & SMRL(image) &  11.33 & 43.5 & 0.30 &-0.2045& 0.7428       \\ \cline{2-7} 
                            & RAMRL & 12.47 & 41.9 & \textbf{0.00} & -0.0509 & \textbf{2.0010} \\ \hline
\end{tabular}

\end{adjustbox}
\caption{Robustness comparison between conventional Single-modality RL (SMRL) and the proposed RAMRL under various noisy observation conditions.}
\vspace{-1em}
\label{tab:robustnessComparision}

\end{table*}

\subsection{Experimental Results}\label{sec:training}

\textbf{Training Performance Comparison.}
We first compare the training performance of the reinforcement learning algorithm when the single modality is considered versus the proposed RAMRL algorithm. Fig. \ref{fig:trainingleModality} shows the normalized episodic returns for a single modality and the proposed RAMRL model trained for $500,000$ steps at $0.0003$ learning rate. The RAMRL achieves a faster convergence rate compared to the single modality models, which makes RAMRL more sample-efficient. The efficacy of the policy is determined once it surpasses the rewards of the Krauss model when all its safety checks are off. We observe the performance of the Image-based model surpasses the BSM model. This is because the BSM data contains partial state information when compared to image data. BSM model using this partially observable MDP achieves a lower overall performance. However, the BSM model has a higher overall driving comfort. Our proposed RAMRL hence combines the advantages of these single modality models while adding robustness to noise. 

\vspace{0.5em}
\textbf{Effectiveness Evaluation.}
We consider several critical metrics regarding driving safety, including the average number of collisions, the driving comfort by average change in acceleration and the traffic flow with the velocity of vehicles after merge.  
First,  we compare the model with the Krauss car-following model. Table~\ref{tab:modalityTesting} compares the performance of the different modalities. 
The single modality models outperforms the rule based models in terms of merge time, normalized episodic rewards. 
The merge time is $34\%$ lower when using single modality versus the rule based methods. The multi-modality and proposed multi-modality with noise model has $19\%$ lower merge time.
However, there are around $10\%$ when using the single modality models. Both single modality BSM and multi-modality models have similar comfort level when compared to the baseline model with a $5-7\%$ change in rate of change of acceleration. Additionally, we compare the resulted traffic congestion or traffic flow of the vehicles on main road due to merging vehicle. We observe that the overall traffic flow defined by the average velocity in $m/s$ remains relatively same when we use the any type of modality. Thus using RL techniques we can reduce the merge time without compromising driving traffic flow or driving comfort.  However, The robust multi-modal approaches outperforms the single modality model in terms of safety. This increases the reliability in noisy environments as shown in Table~\ref{tab:robustnessComparision}.  

\vspace{0.2em}
\textbf{Robustness Evaluation. }
We evaluate the robustness of our approach on different levels of noisy observations for both the BSM and image data. We add 50\%, 25\%, and 10\% of the noise described in Section~\ref{sec:RLPreliminaries}. The BSM data in noisy observation compromises the safety with collisions reaching 30\%. On the other hand, the multi-modality model achieves a low collision rate even in imperfect conditions. Further, when we compare the traffic flow, there is a slight increase of 1.3\% in average velocity as the noise in the observation increases. Furthermore, the driving comfort is the highest when using single modality BSM . Finally, in noisy observation, the total episodic reward decreases due to the inaccuracy in the observations without compromising safety. Using the RAMRL approach, there is a significant increase in safety even in imperfect observations while maintaining traffic efficiency and driving comfort.

\begin{figure}[!tb]
\vspace{-1em}
\centerline{\includegraphics[width=0.4\textwidth]{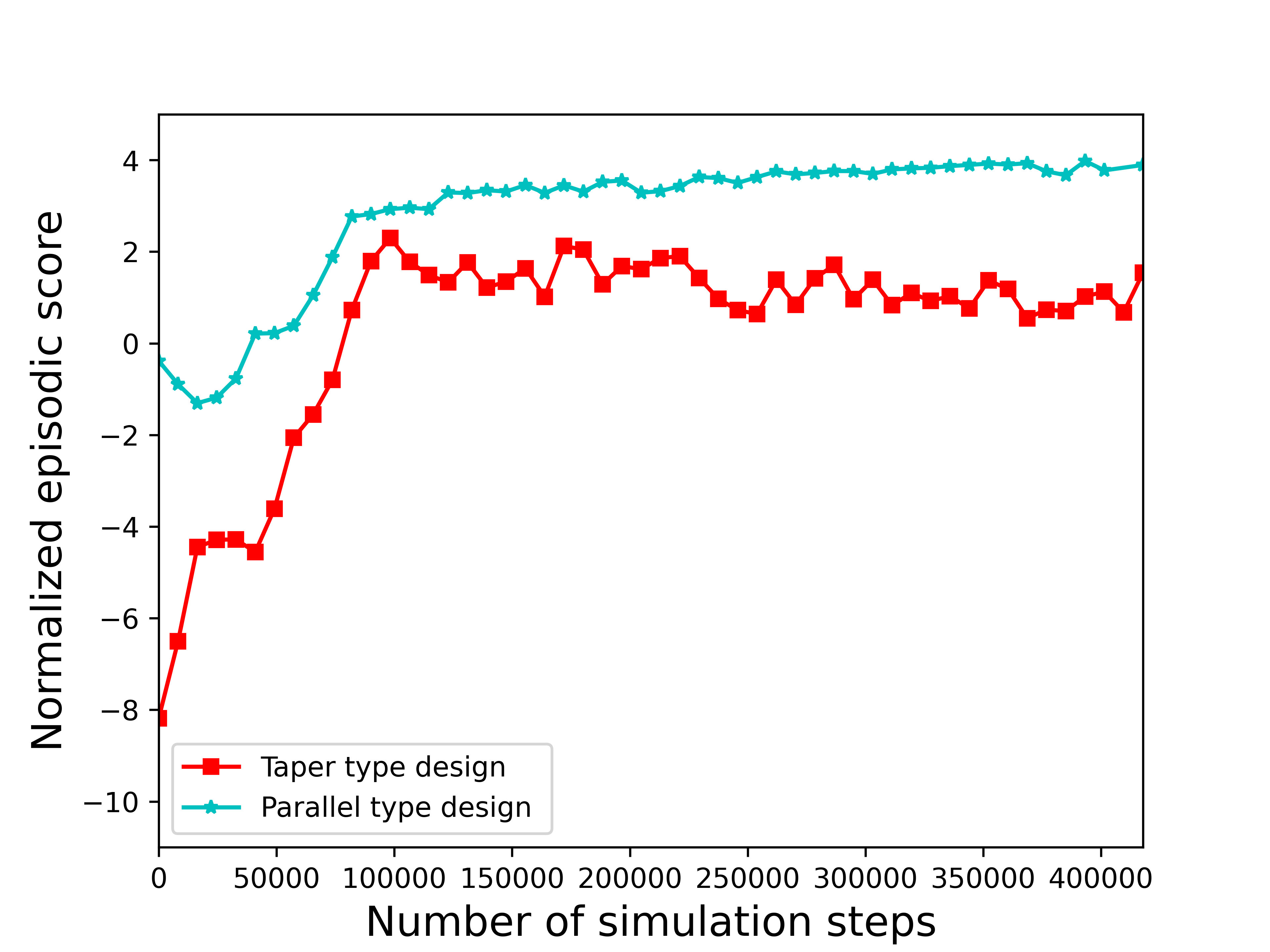}}
\caption{Training curves for taper-type and parallel-type merging scenarios based on RAMRL model.} \label{fig:parallelEpisodicRewards}
\vspace{-1em}
\end{figure}

\vspace{0.5em}
\textbf{Generalization Performance.}
The proposed RAMRL algorithm can be generalized to different types of intersection design. To demonstrate this we create a parallel type design, which has an additional lane for accelerating merging lane.  We compare the parallel type design with the taper type design which we have extensively tested in this paper. Fig. ~\ref{fig:parallelEpisodicRewards} demonstrates training performance of two topologies in Fig. ~\ref{fig:RampToplogies}.  We observe that the taper type design earlier to train for the RL algorithm, which validates the merge type design preferred by AASHTO~\cite{aashto2001policy}. Higher episodic rewards in the parallel type design ensures driving comfort, safety and lower merge time.


\section{Conclusion}
This paper presented a Robust Augmented Multi-modality RL (RAMRL) approach for robust on-ramp merging. The on-ramp merging problem was first formulated as an MDP with a multi-stage reward function to ensure merging safety while maintaining traffic efficiency and driving comfort. 
Besides, RAMRL simultaneously utilizes image and BSM information as the multi-modal observation and outputs the merging maneuvers based on the PPO. The robustness of RAMRL is further improved by the augmented observations.
Extensive experiments were conducted 
to verify the effectiveness 
of the RAMRL approach over baseline approaches, 
especially under imperfect observations, e.g., even 50\% noise for both BSM and image data. 
Lastly, we demonstrated that our approach can be generalized to various on-ramp topologies, which USDOT prefers. 
Our future research will investigate adaptive weights on different modalities based on the amount of perturbation.

\bibliographystyle{IEEEtran}
\bibliography{main}

\begin{thebibliography}{10}
\providecommand{\url}[1]{#1}
\csname url@samestyle\endcsname
\providecommand{\newblock}{\relax}
\providecommand{\bibinfo}[2]{#2}
\providecommand{\BIBentrySTDinterwordspacing}{\spaceskip=0pt\relax}
\providecommand{\BIBentryALTinterwordstretchfactor}{4}
\providecommand{\BIBentryALTinterwordspacing}{\spaceskip=\fontdimen2\font plus
\BIBentryALTinterwordstretchfactor\fontdimen3\font minus
  \fontdimen4\font\relax}
\providecommand{\BIBforeignlanguage}[2]{{%
\expandafter\ifx\csname l@#1\endcsname\relax
\typeout{** WARNING: IEEEtran.bst: No hyphenation pattern has been}%
\typeout{** loaded for the language `#1'. Using the pattern for}%
\typeout{** the default language instead.}%
\else
\language=\csname l@#1\endcsname
\fi
#2}}
\providecommand{\BIBdecl}{\relax}
\BIBdecl

\bibitem{USDOT2015traffic}
\BIBentryALTinterwordspacing
U.~D. of~Transportation Federal Highway~Administration, ``Traffic safety
  facts,'' 2015. [Online]. Available:
  \url{https://crashstats.nhtsa.dot.gov/Api/Public/ViewPublication/812115}
\BIBentrySTDinterwordspacing

\bibitem{zhang2019machine}
L.~Zhang, L.~Yan, Y.~Fang, X.~Fang, and X.~Huang, ``A machine learning-based
  defensive alerting system against reckless driving in vehicular networks,''
  \emph{IEEE Transactions on Vehicular Technology}, vol.~68, no.~12, pp.
  12\,227--12\,238, 2019.

\bibitem{zhao2020sim}
W.~Zhao, J.~P. Queralta, and T.~Westerlund, ``Sim-to-real transfer in deep
  reinforcement learning for robotics: a survey,'' in \emph{2020 IEEE Symposium
  Series on Computational Intelligence (SSCI)}.\hskip 1em plus 0.5em minus
  0.4em\relax IEEE, 2020, pp. 737--744.

\bibitem{schulman2017proximal}
J.~Schulman, F.~Wolski, P.~Dhariwal, A.~Radford, and O.~Klimov, ``Proximal
  policy optimization algorithms,'' \emph{arXiv preprint arXiv:1707.06347},
  2017.

\bibitem{triest2020learning}
S.~Triest, A.~Villaflor, and J.~M. Dolan, ``Learning highway ramp merging via
  reinforcement learning with temporally-extended actions,'' in \emph{2020 IEEE
  Intelligent Vehicles Symposium (IV)}.\hskip 1em plus 0.5em minus 0.4em\relax
  IEEE, 2020, pp. 1595--1600.

\bibitem{lin2020anti}
Y.~Lin, J.~McPhee, and N.~L. Azad, ``Anti-jerk on-ramp merging using deep
  reinforcement learning,'' in \emph{2020 IEEE Intelligent Vehicles Symposium
  (IV)}.\hskip 1em plus 0.5em minus 0.4em\relax IEEE, 2020, pp. 7--14.

\bibitem{liu2021efficient}
J.~Liu, W.~Zhao, and C.~Xu, ``An efficient on-ramp merging strategy for
  connected and automated vehicles in multi-lane traffic,'' \emph{IEEE
  Transactions on Intelligent Transportation Systems}, vol.~23, no.~6, pp.
  5056--5067, 2021.

\bibitem{xiao2020multimodal}
Y.~Xiao, F.~Codevilla, A.~Gurram, O.~Urfalioglu, and A.~M. L{\'o}pez,
  ``Multimodal end-to-end autonomous driving,'' \emph{IEEE Transactions on
  Intelligent Transportation Systems}, 2020.

\bibitem{chaturvedi2022pay}
S.~S. Chaturvedi, L.~Zhang, and X.~Yuan, ``Pay" attention" to adverse weather:
  Weather-aware attention-based object detection,'' \emph{arXiv preprint
  arXiv:2204.10803}, 2022.

\bibitem{laskin2020reinforcement}
M.~Laskin, K.~Lee, A.~Stooke, L.~Pinto, P.~Abbeel, and A.~Srinivas,
  ``Reinforcement learning with augmented data,'' \emph{Advances in neural
  information processing systems}, vol.~33, pp. 19\,884--19\,895, 2020.

\bibitem{Ding2020Rule}
J.~Ding, L.~Li, H.~Peng, and Y.~Zhang, ``A rule-based cooperative merging
  strategy for connected and automated vehicles,'' \emph{IEEE Transactions on
  Intelligent Transportation Systems}, vol.~21, no.~8, pp. 3436--3446, 2020.

\bibitem{Subraveti2018Rule}
H.~H. S.~N. Subraveti, V.~L. Knoop, and B.~van Arem, ``Rule based control for
  merges: Assessment and case study,'' in \emph{2018 21st International
  Conference on Intelligent Transportation Systems (ITSC)}, 2018, pp.
  3006--3013.

\bibitem{NGSIMDataset}
\BIBentryALTinterwordspacing
U.~S.~D. of~Transportation Federal Highway~Administration, ``Next generation
  simulation (ngsim) vehicle trajectories and supporting data. [dataset],''
  2016. [Online]. Available:
  \url{https://data.transportation.gov/Automobiles/Next-Generation-Simulation-NGSIM-Vehicle-Trajector/8ect-6jqj}
\BIBentrySTDinterwordspacing

\bibitem{wang2017modeling}
E.-g. Wang, J.~Sun, S.~Jiang, and F.~Li, ``Modeling the various merging
  behaviors at expressway on-ramp bottlenecks using support vector machine
  models,'' \emph{Transportation research procedia}, vol.~25, pp. 1327--1341,
  2017.

\bibitem{sun2018modeling}
J.~Sun, K.~Zuo, S.~Jiang, and Z.~Zheng, ``Modeling and predicting stochastic
  merging behaviors at freeway on-ramp bottlenecks,'' \emph{Journal of Advanced
  Transportation}, vol. 2018, 2018.

\bibitem{shao2019survey}
K.~Shao, Z.~Tang, Y.~Zhu, N.~Li, and D.~Zhao, ``A survey of deep reinforcement
  learning in video games,'' \emph{arXiv preprint arXiv:1912.10944}, 2019.

\bibitem{lee2021deep}
B.~Lee, V.~Saj, M.~Benedict, and D.~Kalathil, ``A deep reinforcement learning
  control strategy for vision-based ship landing of vertical flight aircraft,''
  in \emph{AIAA AVIATION 2021 FORUM}, 2021, p. 3218.

\bibitem{kiran2021deep}
B.~R. Kiran, I.~Sobh, V.~Talpaert, P.~Mannion, A.~A. Al~Sallab, S.~Yogamani,
  and P.~P{\'e}rez, ``Deep reinforcement learning for autonomous driving: A
  survey,'' \emph{IEEE Transactions on Intelligent Transportation Systems},
  2021.

\bibitem{koepke1993ramp}
F.~J. Koepke, ``Ramp exit/entrance design--taper versus parallel and critical
  dimensions,'' \emph{Transportation Research Record}, no. 1385, 1993.

\bibitem{BSMmandate}
\BIBentryALTinterwordspacing
NHTSA. (2016) Federal motor vehicle safety standards; v2v communications.
  [Online]. Available:
  \url{https://www.nhtsa.gov/sites/nhtsa.gov/files/documents/v2v\_nprm\_web\_version.pdf}
\BIBentrySTDinterwordspacing

\bibitem{wang2021interpretable}
H.~Wang, H.~Gao, S.~Yuan, H.~Zhao, K.~Wang, X.~Wang, K.~Li, and D.~Li,
  ``Interpretable decision-making for autonomous vehicles at highway on-ramps
  with latent space reinforcement learning,'' \emph{IEEE Transactions on
  Vehicular Technology}, vol.~70, no.~9, pp. 8707--8719, 2021.

\bibitem{nuthalapati2016reliability}
N.~Nuthalapati, V.~S. Koganti, and L.~K. Galla, ``Reliability of warning light
  device applications,'' in \emph{2016 International Conference on Control,
  Instrumentation, Communication and Computational Technologies
  (ICCICCT)}.\hskip 1em plus 0.5em minus 0.4em\relax IEEE, 2016, pp. 425--430.

\bibitem{nishitani2020deep}
I.~Nishitani, H.~Yang, R.~Guo, S.~Keshavamurthy, and K.~Oguchi, ``Deep merging:
  Vehicle merging controller based on deep reinforcement learning with
  embedding network,'' in \emph{2020 IEEE International Conference on Robotics
  and Automation (ICRA)}.\hskip 1em plus 0.5em minus 0.4em\relax IEEE, 2020,
  pp. 216--221.

\bibitem{cowlagi2021Risk}
R.~V. Cowlagi, R.~C. Debski, and A.~M. Wyglinski, ``Risk quantification for
  automated driving using information from v2v basic safety messages,'' in
  \emph{2021 IEEE 93rd Vehicular Technology Conference (VTC2021-Spring)}, 2021,
  pp. 1--5.

\bibitem{wen2019high}
T.~Wen, Z.~Xiao, K.~Jiang, M.~Yang, K.~Li, and D.~Yang, ``High precision target
  positioning method for rsu in cooperative perception,'' in \emph{2019 IEEE
  21st International Workshop on Multimedia Signal Processing (MMSP)}, 2019,
  pp. 1--6.

\bibitem{yang2019deep}
Y.~Yang, ``A deep reinforcement learning architecture for multi-stage optimal
  control,'' \emph{arXiv preprint arXiv:1911.10684}, 2019.

\bibitem{hu2020end}
H.~Hu, Z.~Lu, Q.~Wang, and C.~Zheng, ``End-to-end automated lane-change
  maneuvering considering driving style using a deep deterministic policy
  gradient algorithm,'' \emph{Sensors}, vol.~20, no.~18, p. 5443, 2020.

\bibitem{sutton2018reinforcement}
R.~S. Sutton and A.~G. Barto, \emph{Reinforcement learning: An
  introduction}.\hskip 1em plus 0.5em minus 0.4em\relax MIT press, 2018.

\bibitem{deepRL-2020}
H.~D.~Z. Ding, S.~Zhang, H.~Yuan, H.~Zhang, J.~Zhang, Y.~Huang, T.~Yu,
  H.~Zhang, and R.~Huang, \emph{Deep Reinforcement Learning: Fundamentals,
  Research, and Applications}, S.~Z. Hao~Dong, Zihan~Ding, Ed.\hskip 1em plus
  0.5em minus 0.4em\relax Springer Nature, 2020,
  {http://www.deepreinforcementlearningbook.org}.

\bibitem{schulman2015trust}
J.~Schulman, S.~Levine, P.~Abbeel, M.~Jordan, and P.~Moritz, ``Trust region
  policy optimization,'' in \emph{International conference on machine
  learning}.\hskip 1em plus 0.5em minus 0.4em\relax PMLR, 2015, pp. 1889--1897.

\bibitem{raffin2021stable}
A.~Raffin, A.~Hill, A.~Gleave, A.~Kanervisto, M.~Ernestus, and N.~Dormann,
  ``Stable-baselines3: Reliable reinforcement learning implementations,''
  \emph{Journal of Machine Learning Research}, 2021.

\bibitem{lopez2018microscopic}
P.~A. Lopez, M.~Behrisch, L.~Bieker-Walz, J.~Erdmann, Y.-P. Fl{\"o}tter{\"o}d,
  R.~Hilbrich, L.~L{\"u}cken, J.~Rummel, P.~Wagner, and E.~Wie{\ss}ner,
  ``Microscopic traffic simulation using sumo,'' in \emph{2018 21st
  international conference on intelligent transportation systems (ITSC)}.\hskip
  1em plus 0.5em minus 0.4em\relax IEEE, 2018, pp. 2575--2582.

\bibitem{wu2017flow}
C.~Wu, A.~Kreidieh, K.~Parvate, E.~Vinitsky, and A.~M. Bayen, ``Flow: A modular
  learning framework for autonomy in traffic,'' \emph{arXiv preprint
  arXiv:1710.05465}, 2017.

\bibitem{Lucas2019sumorl}
L.~N. Alegre, ``{SUMO-RL},'' \url{https://github.com/LucasAlegre/sumo-rl},
  2019.

\bibitem{Zaid2019SAE}
\BIBentryALTinterwordspacing
F.~Ahmed-Zaid, ``Sae standard overview of basic safety message (bsm),'' 2019.
  [Online]. Available:
  \url{https://www.tomesoftware.com/wp-content/uploads/2019/08/5-2019-B2V-Workshop-Detroit-Farid-Ahmed-Zaid-BSM-Messages.pdf}
\BIBentrySTDinterwordspacing

\bibitem{krauss1998microscopic}
S.~Krau{\ss}, ``Microscopic modeling of traffic flow: Investigation of
  collision free vehicle dynamics,'' 1998.

\bibitem{aashto2001policy}
A.~AASHTO, ``Policy on geometric design of highways and streets,''
  \emph{American Association of State Highway and Transportation Officials,
  Washington, DC}, vol.~1, no. 990, p. 158, 2001.

\end{thebibliography}
\end{document}